\RequirePackage{amsmath}
\documentclass[runningheads]{llncs}
\usepackage{makeidx}         
\usepackage{graphicx}        
\usepackage{multicol}        
\usepackage[bottom]{footmisc}

\usepackage[utf8]{inputenc}

\usepackage{newtxtext}       %
\usepackage{newtxmath}       
\usepackage{tabularx}
\usepackage{multirow}

\usepackage{bm}

\newcommand{\R}{{\mathbb R}}

\newcommand{\be}{\begin{equation}}
\newcommand{\ee}{\end{equation}}
\newcommand{\ba}{\begin{array}}
\newcommand{\ea}{\end{array}}
\newcommand{\baa}{\left[\begin{array}}
\newcommand{\eaa}{\end{array}\right]}

\newcommand{\beqa}{\begin{eqnarray}}
\newcommand{\eeqa}{\end{eqnarray}}
\newcommand{\bt}{\begin{tabular}}
\newcommand{\et}{\end{tabular}}

\begin{document}

\title{Quick survey of graph-based fraud detection methods \thanks{This work was supported by BRD Groupe Societe Generale through Data  Science Research Fellowships of 2019. Andra Băltoiu was also supported by the Operational Programme Human Capital of the Ministry of European Funds through the Financial Agreement 51675/09.07.2019, SMIS code 125125.}}

\author{Paul Irofti\inst{1,2}\and 
Andrei P\u{a}trașcu\inst{1} \and 
Andra B\u{a}ltoiu\inst{2}}

\institute{Department of Computer Science, University of Bucharest, Romania \and
The Research Institute University of Bucharest (ICUB), Romania\\
}

\maketitle

\begin{abstract}In general, anomaly detection is the problem of distinguishing between normal data samples with well defined patterns or signatures and those that do not conform to the expected profiles. Financial transactions, customer reviews, social media posts are all characterized by relational information.
In these networks, fraudulent behaviour may appear as a distinctive graph edge, such as spam message, a node or a larger subgraph structure, such as when a group of clients engage in money laundering schemes. 
Most commonly, these networks are represented as attributed graphs, with numerical features complementing relational information. We present a survey on anomaly detection techniques used for fraud detection that exploit both the graph structure underlying the data and the contextual information contained in the attributes.

\end{abstract}

\section{Introduction}
\label{Sec:intro}

The value of fraudulent transactions in the EU was about 1.8 billion euro in 2016 only for card fraud \cite{EuFr:09}.
Many other financial criminal activities, like money laundering (e.g. through multiple or over-invoicing), corruption-related transfers, VAT evasion, identity theft, etc., affect individuals, banks and state activities like tax collection.
As the number of transactions increases and criminal behavior becomes more sophisticated, fraud detection requires more attention and time from human experts employed by banks or state authorities.
The need of performant automatic tools for at least selecting the most likely fraudulent activities, but aiming also to detect new types of ill-intentioned activities, is imperative.
This is by no means valid only in the case of financial crimes. Insurance frauds, e-commerce or social networks misconducts such as fake reviews are other examples of domains where securing activities are critical.

Money transactions (payments, transfers, cash withdrawals, etc.) can be described by a vector of characteristics.
However, treating transactions as independent vectors is an over-simplification, due to the intricacies of many types of criminal behavior.
It is much more adequate to treat the transactions in their natural form, that of a {\em graph} whose nodes are the financial entities (individuals, firms, banks) and whose edges are transactions data (amount, time, payment mode, etc.). A directed edge between two nodes illustrates that there is a money transfer in the respective direction, where the weight on the edge is the transferred amount.
Graphs allow to model inter-dependencies, capture the relational nature of transactions and are also a more robust tool, as fraudsters usually do not have a global view of the graph.
Some frauds, in other words, imply a particular scheme of relationships that can be revealed only by looking at the underlying graph structure. 

Most of the times, the data consists of an attributed graph, with attributes either on the nodes, edges or both. 
Fraud detection methods thus need to integrate both relational knowledge, accessible from the network structure, and the numeric features that describe graph components. 
Depending on the application, anomalies can be considered at the level of a node, an edge, or a subgraph.

We survey several types of topological features that can be exploited in order to differentiate legitimate graph entities from fraudulent ones and examine how relational information can be used in tandem with other numerical data sources.

Whenever applicable, we also focus on solutions that take into account the evolution of a graph.
The need for including the temporal dimension is two-fold. 
On one hand, in some cases it is the temporal pattern that defines the anomaly. 
Several types of frauds, including credit card frauds, network attacks and spam, involve some sort of high-frequency activity.
It is not the activity per se that gives away its fraudulent nature, but the fact that it occurs often or in a specific pattern. 

On the other hand, there is the pressing need to react as fast as possible to the occurrence of a misconduct.
Fraudulent event detection therefore requires temporal representation.  

Tabel \ref{tbl:summary} presents of a summary of the reviewed papers.
Attributed and non-attrbuted graphs are both considered. Literature is substantially biased towards non-attributed and node-attributed graphs, since edge attributes can often be transfered either to nodes or recast as weights. The second criteria is anomaly localization, namely the envisaged graph entity: node, edge or different subgraph structures (sometimes referenced as group of nodes). We then focus on the type of learning method, since numerous applications require lack of supervision. Finaly, we mention the application domain, focusing on the principal performed tests, where multiple applications are discussed. 

\begin{table}[t] \caption{Review summary}
	\makebox[\textwidth][c]{
		\begin{tabular}{|c|c|c|c|c|}
			\hline
			\textbf{Attributes} & \textbf{Anomaly localization} & \textbf{Learning Method} & \textbf{Dataset} & \textbf{Reference} \\
			\hline
			\hline
			\multirow{4}{*}{Edge} & Edge & Semi-supervised & Financial & \cite{CaoMao:17} \\
			\cline{2-5}
	 		& Node & Unsupervised & Financial & \cite{ShaBeu:16} \\
			\cline{2-5}
			& \multirow{2}{*}{Subgraph} & Unsupervised & Reviews & \cite{DhaGan:19} \\
			\cline{3-5}
			& & Supervised & Financial & \cite{SavWan:16} \\
			\hline	
			
			Edge $\&$ Node & Subgraph & Unsupervised & Financial & \cite{QiuCen18:} \\
			\hline
	
			\multirow{6}{*}{Node} & \multirow{2}{*}{Node} & \multirow{2}{*}{Unsupervised} & Sensor & \cite{VenTha:15} \\
			\cline{4-5}
			 &  &  & Financial & \cite{LiHar:17}, \cite{LiuHua:17}, \cite{PenMin:18} \\
			\cline{2-5}
			& \multirow{4}{*}{Subgraph} & \multirow{4}{*}{Unsupervised} & Reviews & \cite{WanZho:18} \\
			\cline{4-5}
			& & & Social & \cite{Akoglu12_PICS} \\
			\cline{4-5}
			& & & Messaging & \cite{Huang17_JointWN} \\
			\cline{4-5}
			& & & Citation & \cite{LiSun:14}, \cite{MilBea:15}, \cite{PerAko:16} \\
			\hline	
			\multirow{13}{*}{Non-Attributed} & Edge $\&$ Node & Supervised & Social, Messaging & \cite{YuChe:18} \\
			\cline{2-5}
			& \multirow{5}{*}{Node} & \multirow{3}{*}{Unsupervised} & Reviews & \cite{GaoDu:13}, \cite{LiuHoo:17} \\
			\cline{4-5}
			& & & Financial & \cite{ColRem:17}, \cite{Pandit07_netprobe}, \cite{WanYua:20} \\
			\cline{4-5}
			& & & Social & \cite{Sen:18} \\
			\cline{3-5}
			& & Semi-supervised & Reviews & \cite{Yuan17_deepspectrum} \\
			\cline{3-5}
			& & Supervised & Network & \cite{HenKei:2011} \\
			\cline{2-5}
			& \multirow{7}{*}{Subgraph} & \multirow{5}{*}{Unsupervised} & Reviews & \cite{BanLiu:19} \\
			\cline{4-5}
			& & & Network & \cite{VelEbe:17} \\
			\cline{4-5}
			& & & Financial & \cite{EllCuc:18}, \cite{BPI20_ifac}, \cite{LiXio:10}, \cite{MilBea:15}, \cite{ZhaZho:17} \\
			\cline{4-5}
			& & & Social & \cite{AkoMcG:10}, \cite{JiaCui:16}, \cite{MilBer:11a}, \cite{Ying11_socialspectrum} \\
			\cline{3-5}
			& & \multirow{2}{*}{Supervised} & Messaging & \cite{CheHen:12} \\
			\cline{4-5}
			& & & Social & \cite{JiaMen:14} \\
			\hline			 
		\end{tabular}
	}\label{tbl:summary}
\end{table}  

In the remainder of the paper, we first we survey literature sectors which orbit around the idea of detecting anomalies in attributed networks through promotion of particular subgraphs having unusual structure, such as high density or specific connectivity patterns like rings, cliques or heavy paths. These methods examine the local patterns of the graph, by working with neighborhoods. 

We then focus on methods that consider the graph structure from the perspective of each node and finish our survey by looking at methods that create hybrid methods for integrating topological information with network attributes.

\section{Related work}
\label{Sec:related}

Several surveys exist that tackle similar topics.  
Dating 2010, \cite{FleVay:10} focuses on fraud detection techniques from an audit perspective, however it does not cover only machine learning approaches.
A classification-oriented survey on financial fraud identification can be found in \cite{NgaHu:11}, yet it does not review solutions for graphs.
Accuracy and specificity summaries obtained by various machine learning methods on different types of financial frauds can be found in \cite{WesJar:15}.
The survey in \cite{PhuCli:13} presents supervised and unsupervised methods and have a dedicated section to similar applications. 
Credit card fraud detection methods are reviewed in \cite{DelHuss:09} and more recently in \cite{SorZoj:16}.

Fraud detection can be cast as an anomaly detection (AD) task, since frauds are rare events that distinguish themselves from normal behavior. 
As such, methods for anomaly identification on graphs represent a great pool of solutions to the problem of fraud detection. 
Nonetheless, it must be noted that in the absence of an application-driven definition of a graph anomaly, some general methods can be ineffective for our task.
A comprehensive survey on graph anomaly detection can be found in \cite{AkoTon:14}. 

The survey defines structure-based methods as those which seek information in the characteristics of the graph to define (ab)normality: node-level measures such as degree, between nodes measures such as number of common neighbors.
Attributed graphs contain additional information, either on the nodes, thus describing features of the respective entity or on the edges, characterizing the relationship between two entities. 
In this case, structure-based methods are, for the authors of \cite{AkoTon:14}, those methods that look for unusual subgraph patterns. 
The definition is particularly relevant to the fraud detection problem, since frauds are often performed by a group of entities, as a scheme. 
Taken individually, the events making up the scheme may look legitimate, yet they form an anomalous structure.
More on several such structures, successfully identified on dataset of real financial transactions data, however not dealing with attributed graphs, can be found in \cite{EllCuc:18}. 
Community-based methods, on the other hand, assume in their perspective that anomalous nodes do not belong to a community, and are found to be linking communities together.
Note that this definition, however, does not fit common types of financial misconduct.
The authors introduce a third category of relational learning.
Unlike the regular learning paradigm, where independence is assumed between entities, relational learning seeks to incorporate connectivity information, for example by looking at one node's neighbours as well.

\section{Communities}
\label{Sec:community}

Perhaps the most widespread approach when considering connectivity patterns is that of using community information to train a supervised learning system. 
In \cite{SavWan:16}, for example, after decomposing the transaction graph into smaller communities, the authors identify a feature set involving information on network dynamics, party demographics and community structure. They then apply a simple supervised learning method to detect anomalies.

In \cite{PerAko:16} a normality measure is used to quantify the topological quality of communities as well as the focus attributes of communities in attributed graphs. 
In other words, normality quantifies the extent to which a neighborhood is internally consistent and externally separated from its boundary. 
The proposed method discovers a given neighborhood’s latent focus through the unsupervised maximization of its normality. 
The respective communities for which a proper focus cannot be identified receive low score, and are deemed as anomalous.

A modularized anomaly detection hierarchical framework has been developed in \cite{EllCuc:18} to detect static anomalous connected subgraphs, with high average weights. For this purpose, particular community detection strategies are tailored based on 140 features (including Laplacian spectral information) and network comparison tests (such as NetEMD). Then, a classification via random forests or simple sum of individual (feature-based) scores is performed to highlight the anomalous subgraphs.

In directed trading networks, blackhole and volcano patterns represent groups of nodes with inlinks only from the rest of nodes or only outlinks towards the rest nodes, respectively. 
These kinds of patterns, which often have fraudulent nature, are isolated in  \cite{LiXio:10} through pruning (divide-et-impera) schemes based on structural features of blackholes and volcanoes.

In \cite{SavWan:16}, community detection methods are combined with supervised learning for detecting money laundering groups in financial transactions. Community detection begins with extraction of (possibly overlapping) connected components from the transactions attributed multi-graph. Since typical fraudulent communities contain a small number of vertices (less than 150),  the excessively large connected components extracted from the AUSTRAC dataset \cite{latimer1996australia} are further decomposed through a $k-$step neighborhood strategy. This entire process leads to a collection of small communities which are classified through a supervised learning scheme (SVM and random forests).

The authors of~\cite{Ying11_socialspectrum}
start from a clean graph $G$
that has been perturbed by additional anomalous nodes and vertices $N$
resulting in graph $\tilde{G} = G + N$.
The nodes $N$ create a cliques amongst themselves to mimic authentic communities and avoid detection through standard topological based methods.
But, these nodes also have to reach-out to the existing nodes in $G$ for the purpose of fraud or other malign intentions.
The $N$ nodes randomly attack victim nodes from $G$ in order to further avoid detection.
The combination of these techniques is called Random Link Attack (RLA) in ~\cite{Ying11_socialspectrum}.
In this setup,
anomaly detection is performed by taking into account the $k$ largest eigenvalues and their associated eigenvectors.

Within communities, uncommon subgraphs can be mined using structural information, as shown in \cite{BPI20_ifac}. 
The authors propose adaptations of the dictionary learning problem to incorporate connectivity patterns. 
One such adaptation involves imposing that the dictionary atoms express a Laplacian structure, thus creating a dictionary of elementary relational patterns.

Evolutionary networks are considered in \cite{CheHen:12}, where a community detection strategy is used to highlight anomalies based on the temporal quantitative evolution of network communities.

In \cite{QiuCen18:} constrained cycles are detected in dynamic graphs and labeled as fraudulent activities in financial payments system (fake transactions). The authors consider a directed attributed graph with varying edge-structure over time. For each incoming edge between vertices $(u,v)$, efficient algorithms are given to generate all fixed $k-$length cycles between $u$ and $v$. Based on empirical observed issues in the case when high-degree vertices (hot points) are encountered in the generated paths, some indexing procedures are proposed to boost time performance of the brute force depth-first search algorithm. The evaluation data is based on real activity from Alibaba's e-commerce platform, containing both static and dynamic edges, resulting in a graph with $\approx 10^9$ vertices and $\approx 10^9$ edges. Results show a somewhat improved performance, guaranteed by the indexing procedure.

\section{Dense Subgraphs}
\label{Sec:dense}

The intuition behind searching for dense blocks in graphs as signs of anomalies is that some frauds are performed by repetitive activity bursts.
When looking at the graph connectivity, these activities form a dense subgraph that stands out from the sparser normal activity. 
In yet other cases, the density is a consequence of multiple malicious users acting similarly and synchronously, a behavior known as lockstep.

The dense subgraph detection problem is approached in \cite{ZhaZho:17}, where the authors consider a hierarchical framework of subgraph detection, where the initial graph is successively filtered in $k$ steps until a dense cluster results. After modeling this problem as the maximization of a nonconvex quadratic finite-sum (with $k$ term) over integers, several relaxations are applied: $(i)$ ordering in node vectors (equivalent with replacing binary constraint $\{0,1\}$ with convex interval $[0,1]$); $(ii)$ penalization of hierarchical density order such that the $k-$th subgraph is more dense than the subgraph $(k-1)$.
Furthermore, as typical for continuous nonconvex problems, a block-coordinate gradient descent algorithm with Armijo stepsize policy is presented and its convergence to a stationary point of QP model is proved. For numerical evaluation, the authors use the  AMiner co-authorship citation network and a financial bank accounts network. Large density clusters detection is proved along with superiority over existing 2-hierarchies strategies.

The approach in \cite{BanLiu:19} considers tensors for modelling large scale networked data. Starting from the fact that formation of dense blocks is the result of certain entities sharing between two or more entries in the tensor, they construct an Information Sharing Graph (ISG) illustrating these relations. An efficient D-Spot algorithm that detects these dense subgraphs is proposed. Along with some theoretical guarantees for the subgraphs densities generated by the algorithm, the empirical evaluation shows that D-Spot has better accuracy than other tensor-based schemes on synthetic and real datasets such as Amazon, DARPA, Yelp and AirForce.

The work in \cite{JiaCui:16} sets the goal to detect suspicious nodes in a directed graph that have synchronized and abnormal connectivity patterns.
First, a mapping is proposed that embeds the data into a chosen feature space. 
Then, synchronicity and normality measures are introduced. Similarity is computed between the points resulting from the embedding as well as normality of the given data features relative to the rest of the data.
Superior performance (precision, recall, robustness) over well-known state-of-the-art static graph anomaly detection techniques such as Oddball and OutRank is shown on three real world datasets, namely TwitterSG, WeiboJanSG an WeiboNovSG, all of which are complete graphs with billions of edges.

\section{Bipartite Graphs}
\label{Sec:bipartite}

A bipartite graph is a graph that can be split into distinct two subsets such that all edges connect a node from the first subset to one from the second subset.
In some fraud detection problems, a bipartite (sub)graph occurs naturally as a result of scams, or is a convenient way of representing the data. 

The work in \cite{Pandit07_netprobe} identifies two types of identities in auction networks besides honest users: frauds and accomplices.
The latter category supports frauds, but also acts as a camouflage, by adding legitimate activities to the frauds' repertory.
The networks these two types of users create form a bipartite core within the large graph.
The authors then develop a belief propagation algorithm, which infers the identity of a node by evaluating the neighbours.
An adaptation is also provided that efficiently solves the identification problem when the graph structure evolves in time.

Cases when the two classes of nodes involve mutual relations are approached in \cite{LiuHoo:17}, where bipartite graphs are also used to represent data. The fraudulent instances considered in \cite{LiuHoo:17} are assumed to satisfy some given empirically observed traits such as: fraudsters engage as much firepower as possible to boost customer objects, suspicious objects seldom attract non-fraudulent users and fraudulent attacks are well represented by bursts of activity. Further, they detect fraudulent blocks corresponding to both vertices sets in the bipartite graph and formulate a metric that measures to what extent a given block obeys the fixed traits. By maximizing this metric over the entire data, suspicious block are labeled. The experiments show that HoloScope achieves significant accuracy improvements on synthetic and real data, compared with other fraud detection methods.

More particular bipartite reviewer-product data is considered in \cite{DhaGan:19} and, using unsupervised algorithmic heuristics, the authors aim to find fraudulent groups of reviewers, that typically write  fraud  reviews  to  pro-mote/demote certain products. The designed architecture detects suspicious groups by several coherent behavioral signals of reviewers based on particular quantitative measures such as: reviewer tightness, neighbor tightness and product tightness. 
Experiments on four real-world labeled datasets (including Amazon and Playstore)
show that the DeFrauder algorithm outperforms certain baselines, having $11.35\%$ higher accuracy in group detection.

Further, in \cite{WanZho:18}, a novel fraud detection method is introduced based on deep learning. The data, representable as a bipartite graph is embedded into a latent space such that the representations of the suspicious users in the same fraud block sit as close as possible, while the representations of the normal users are distributed uniformly in the remaining latent space. In this way, the additional density-based detection methods might easily detect the fraud blocks. 
In fact, the deep model from \cite{WanZho:18} involves an autoencoder used to reconstruct the "user" nodes information from the bipartite graph and, at the same time, to ensure that in the low dimensional latent space the anomalous instances are sufficiently similar with respect to a proposed similarity measure.
Thus, the objective function of the minimization problem contains the nonlinear composite terms associated to reconstruction and similarity. 
The tests show that the model is robust in automatically detecting
multiple fraud blocks without predefining the block number, in comparison with other state-of-the-art baselines (which do no rely on deep learning) such as Holoscope, D-cube and others.

\section{Spectral Localization}
\label{Sec:spectral}

As stated in \cite{EllCuc:18}, \emph{spectral localization is the phenomenon in which a large amount of the mass of an eigenvector is placed on a small number of its entries}. 
The methods surveyed in this section use spectral information to find particular subgraphs and thus distinguish anomalies. 
While not all consider the case of attributed graphs and some rely only on topological information, the selection is relevant for the descriptive power of eigenvalues and eigenvectors, which can be subsequently leveraged to include nodes or edge attributes.

Related to Laplacian matrices, the dominant components of their eigenvectors correspond to nodes in the network with special properties, and thus constitute good candidates for the anomaly detection task (see \cite{CucBlo:13,PasCas:16}).
In the series of papers \cite{MilArc:13,MilBer:11a,MilBea:15,MilBli:10b} a set of schemes are developed in order to uncover anomalies using 
spectral features of the modularity matrix. Furthermore, the authors of \cite{MilBea:15} extend these methods to use the $\ell_1$ norm for eigenvectors of sparse Principal Component Analysis (PCA), which performs well at the cost of being more computationally intensive. 
The work in \cite{GaoDu:13} is a cross data source application, that uses a spectral embedding of individuals in different data sources, declaring an anomaly if the embeddings deviate substantially. 

Let $x_j$ be the eigenvector of $\lambda_j$ and $\alpha_j$ and $\beta_j$ be the corresponding elements of the attacker and the victim nodes respectively.

The work in \cite{Ying11_socialspectrum} provides a result that pinpoints the change of each eigenvector element $\alpha_{ij}$ which is directly tied to attacker node $i$ based on its victims from $\beta_{j}$.
Based on this insight attacking nodes can be separated from victim nodes
due to the special distribution of their corresponding eigenvector elements that can be upper bounded thus unveiling a non-random pattern that is not observed in victim nodes.

A similar approach is presented in~\cite{Wu13_spectralAD}
where the attacker nodes are considered subtle subgraphs
with minor impact on the overall spectrum of the general graph (termed the background graph).
The authors motivate the selection of the first largest $k$ eigenvalues by showing that when adding attacker nodes to an existing graph, the added information can only be separated in the first two eigenvectors $x_1$ and $x_2$.
Afterwards the victim and attacker node information is indistinguishable.
With this result they continue with the actual anomaly detection where the two types can be separated in similar fashion: based on the distribution of their associated entries.

Lockstep behavior can also be revealed by spectral properties, as shown in \cite{JiaMen:14}. 
The authors identify different anomalous connectivity patterns by bridging adjacency information with information from the singular vectors.

The authors of~\cite{Yuan17_deepspectrum} use spectral information to train a deep autoencoder and a convolutional neural network based on the RAL attack model and the spectral strategy from~\cite{Ying11_socialspectrum}.
They start with the $k$ largest eigenvectors of the graph $G$ that they arrange as the columns of matrix $M \in \R^{k\times V}$
and
for each vertex $v_j$
they select the corresponding row from $M$.
Then they proceed to compute the mean vector spectral of its $n$-nearest neighbours,
iterating from $1$ to $n$.
The input for the neural network is obtained by concatenating the above vectors into a single large feature vector.

\section{Egonets}

So far, we have surveyed methods that consider a panoptic view of the graph, where topological information is typically conveyed by connectivity matrices such as the adjacency matrix and graph Laplacian or is summarized using different network statistics.
We turn now to alternative ways of exploiting connectivity information, as well as to the case where a single graph has limited descriptive power.

An egonet is the induced subgraph formed by the neighbors of a single node. 
The authors of \cite{AkoMcG:10} have done an extensive research on several real-world networks and found that using carefully extracted, but otherwise intuitive features for describing an egonet, unforeseen normality power laws appear. 
These patterns describe dependencies such as between the number of nodes and edges, between the weights and number of edges, the principal eigenvalue and total edge weight.
The power laws, which have been tested for both uni- and bipartite, weighted and unweighted graphs, encourage the use of various outlier detection metrics. 
A mix of distance-based heuristic and a density-based score is used in \cite{AkoMcG:10}.
The method is used to identify several types of graph anomalies such as (near-)cliques, stars, heavy vicinities and heavy links.

The work in \cite{Sen:18} uses the egonet approach and a thresholding technique to detect anomalous cliques, considering statistical measures of the underlying graph model.
A similar solution uses egonets for fraud detection in e-banking transaction data \cite{WanYua:20}. 
Authors use the Mahalanobis distance metric to label anomalous accounts.

A combination of egonet attributes and node features is used in \cite{HenKei:2011}, together with a set of recursive features.
The latter consist of aggregated means and sums of different network metrics.  
The method thus constructs an abstract characterization of a node that serves in classification and de-anonymization (identity resolution) tasks.

\section{Hybrid Clustering}
\label{Sec:clusters}

This section is concerned with attempts at integrating topological information with the contextual information of the graph attributes in order to go beyond subgraph pattern matching and instead use hybrid measures that aggregate structure data with node or edge descriptions.

In many financial databases, individual isolated fraudulent transactions appear as normal entries if one relies only on local statistical feature analysis. 
However, from a more global perspective, following certain dependencies between transactions leads to a better indicator of anomalous payments.
Inspired by these observations, in \cite{CaoMao:17} the authors seek fraudulent payments in e-commerce transactions networks, based on inter-transaction graph dependencies. The work models the real payment transaction Electronic Arts database as a Heterogenous Information Networks (HIN), which represents an attributed graph with multiple nodes and edge types. The approach is strongly based on several meta-path concepts that are introduced in order to extract semantic relations among multiple transactions (using the paths connecting them) and aggregation of label information. Over baseline methods such as SVM and Random Forests, the proposed HitFraud algorithm provides a boost of 7.93\% in recall and 4.63\% in F-score on the EA data.

An integrated anomaly detection framework for attributed networks is proposed in \cite{LiuHua:17}. A preliminary clustering strategy is presented, which provides the degrees to which attributes are associated to each cluster. Then, a subsequent unsupervised learning procedure is applied based on the representation of the links and data attributes by the set of outcome vectors from the clustering stage. Finally, the abnormal attributes and their corresponding degrees of abnormality are computed on the basis of these representations.

In \cite{Akoglu12_PICS}, clusters are constructed of nodes that have "similar connectivity" 
and exhibit "feature coherence", based on the intuition that clusters are a way to compress the graph. 
As such, the Minimum Description Length (MDL) principle is used to derive a cost function that encodes both the connectivity matrix and the feature matrix.

Edge-attributed networks are considered from an information-theoretic perspective based on Minimum Description Length in \cite{ShaBeu:16} as well. 
The algorithm consists of a combination between an aggregation step of neighborhood attributes information and a clustering step used to provide an abnormality score on each node using the aggregated data. 
MDL is also used in \cite{VelEbe:17}, where normative graph
substructures are identified by taking into account some coverage rules and their quantitative occurrence is established. 
Anomalous substructures are selected from those with least occurrences in the graph. 

In \cite{VenTha:15} certain types of anomalies are detected based on scoring each node (or an entire subgraph) using statistical neighborhood information, such as the distance between the attributes of the node and its neighbors.  A combination of this scoring procedure with a deep autoencoder is also provided. 
Several social network statistical metrics and clustering techniques are used in \cite{ColRem:17} to detect fraud in a factoring company. 

The model from \cite{LiHar:17} defines a normal instance as the one that have a sparse representation on the set of instances and the representation residual is minimized. The network structure has been included in the model through a Laplacian type quadratic penalty.
Furthermore, the model developed in \cite{PenMin:18} selects a
subset of representative instances on the space of attributes that are closely hinged with the network topology based on CUR decomposition, and then measures the normality of each instance via residual analysis.

The authors of \cite{LiSun:14} adopt a finite mixture model to interpret the underlying attribute distributions in a vertex-attributed graph. In order to accommodate graph structure, network entropy regularizers are proposed to control the mixture proportions for each vertex, which facilitates assigning vertices into different mixture components. A deterministic annealing expectation maximization is considered to estimate model parameters.

Following a different approach, other attributed graph clustering methods 
seek to jointly optimize the potential of a cluster to incorporate both topological information and node/edge attributes.
Such an approach is proposed in \cite{Huang17_JointWN} and uses nonnegative matrix  factorization to cluster the graph based on topology and on features at the same time.
The objective function is composed of a term that models structure and a term that captures information on attributes.

A latent attributed network representation is learned in \cite{YuChe:18} by using a number of network walks. The representation is obtained through maintaining the pairwise vertex-distance in the local walks and by hybridizing it with the hidden layer of deep auto-encoder, such that the resultant embedding is guaranteed to faithfully reconstruct the original network. Then, a dynamic clustering model is used to flag anomalous vertices or edges based on the learned vertex or edge representations. Moreover, leveraging a reservoir sampling strategy, any dynamic network change induces only modest updates on the learned representations. 

\section{Conclusions}

From transaction attributes to relational structures and encoding information on the time varying properties of a graph, we searched for solutions that promote the aggregation and integration of multiple sources of data.
The review focuses on different perspectives on graph structures that enable the identification of anomalous entities: edges, nodes or subgraphs that are manifest suspicious behaviour.

\bibliographystyle{plain}
\bibliography{review}

\end{document}